\documentclass[a4paper,11pt]{article}
\usepackage{fullpage}
\usepackage{makeidx}
\usepackage{colordvi}
\usepackage[T1]{fontenc}
\usepackage{verbatim}
\usepackage[dvips]{graphicx}
\usepackage{calc}

\makeindex

\begin{document}
\date{}

\title{The Semantics of Kalah Game }
\author{Kaninda Musumbu\\
 LaBRI~ (UMR 5800 du CNRS),\\
Universit\'e Bordeaux I, France \\
351, cours de la Lib\'eration, F-33.405 TALENCE Cedex,\\
e-mail: musumbu@labri.fr\\
}

\maketitle

\thispagestyle{empty}
\pagestyle{empty}

\begin{abstract} {\em                                                           The present work consisted in developing a plateau game. 
There are the traditional ones (monopoly, cluedo, ect.) but those 
which interest  us  leave less place at the chance (luck) than to the 
strategy such that the chess game.  Kallah is an old African game, its
 rules are simple but  the strategies to be used are very complex to 
implement. Of course, they are based on a strongly mathematical basis as in 
the film  "Rain-Man" where one can see that gambling can be payed with 
strategies based on mathematical theories. The Artificial Intelligence gives 
the possibility "of thinking"  to a machine and, therefore, 
allows it to make decisions. In our work, 
we use it to give the means to the computer choosing its best movement.
}
\end{abstract} 

\section{Introduction}
The present work consisted in developing a plateau game.   
There are the traditional ones (monopoly, cluedo, long course...) but those 
which interest  us  leave less place to random than to the strategy. 
There are the plays of simulation like: 
\begin{itemize}
 \item "The age of the rebirth". Historical reconstitution from 750 to 1350
 (of the Middle Ages to the rebirth) highly strategic Play, basing itself on
 the commercial conquests (and not soldiers). The play advances thanks to 
discoveries, inventions, etc. 
\item "Risk". It is a military play of strategy, 
consisting in conquering the world. The plate represents the chart of the 
world. Each team must fight, or link itself to remove the opposing armies. 
\item "Dune". Who doesn't know the Dune planet, with his spice, vital for all 
the protagonists? Find the environment of Franck Herbbert, while using of 
diplomacy, strategy and bluff to be the Master of spice. 
\end{itemize} 
One  finds also sets of rules simple but "effective". In which we can integrate the { \em Kalah game}. These rules are simple but the strategies to be used 
are very complex to implement. Of course, they  are based on a strongly mathematical basis.with calculations of the differences of pawns, empty holes... as in the film
 "Rain-Man" where we can see that gambling can be played with strategies 
based on mathematical theories. The { \em Kalah } game  is one of oldest
 African plays. 
It is a mathematical play and the most complex versions can be  compared  with the  chess game. \\
The play of { \em Kalah } is 
appeared as table, composed of two lines of six holes plus two called 
"special" holes { \em Kalahs }. At the beginning of the game, each hole 
contains six pawns and the { \em Kalahs } are empty. The game's goal is
 to collect more than half of the total number of pawns in its { \em Kalah }. 
We have developed the { \em Kalah} game  on the computer with four 
modes of play: 
\begin{itemize}
 \item Two players on the same machine. 
\item a player counters the computer. 
\item Two players in network. 
\item Two computers. 
\end{itemize} 
Paragraph 2 is a general presentation of work with a recall of rules of the 
game. In paragraph 3, we state  the problem posed by the taking into 
account  the modification of the apron following a movement of one of the 
players. In paragraph 4, we explain  the problem arising from the computing 
time necessary to make play the computer and the solutions brought to really
 make the play interactive and acceptable the latency.

\section{General Presentation}
For our implementation we decide to put two lines made up of six holes plus 
two "special" holes called { \em Kalahs }. Each hole contains six pawns at 
the beginning of the play and the { \em Kalahs } are empty. The { \em Kalahs }
 are used to store the pawns collected by each player. The player can choose 
between four modes of play: two computers, a player counters the computer, 
two players on the same machine, two players in network. The course of the 
play can be followed using the messages posted to the medium of the screen 
after each modification made on the table 

\section{ Functionalities of Game Engine }
This module is used for the direct interaction with all the structures of data
 necessary for storage and modifications of the states of the play. It manages
 four modes different of game: 
\begin{itemize} 
\item Two players on the same machine. With each turn, the computer lets to 
the user click on one of his  holes. Once the choice makes the engine carries 
out the movement( the distribution of pawns) on the structures of definite 
data and informs with the interface to refresh the screen. 
\item A player counters the computer. The human player always takes the hand 
at the beginning of the play. He chooses a hole, the engine will carry out this movement and will give the  hand to the computer which will make its movements 
of continuation with a direct interaction with the Strategies module. 
Once all the finished movements,  the graphic interface will start to post the 
movement of human then all the movements of the computers for this blow. 
\item Two players in network. The course of the play is exactly the same one 
as for two players on the same  machine, except that the engine with each 
movement sending  on the distant machine the movement made by the local 
machine in order to post on  two sides the same apron of play. 
\item Two computers. For each computer the Strategy module will return the 
best hole for the state of the current apron by using the strategy and the 
level of currents difficulty. With this hole the engine will make the movement 
and the storage of the apron and will pass the hand to the other computer or 
will remain on itself if it replays. 
\end{itemize}

This module uses in the same way the GameLogic class.
 This class contains all information concerning the rules of the game 
implemented (including the determination of the winner). The structures used 
most significant are for example: The BoardState class which represents an 
apron of play storing all the values of the pawns in the holes and the 
{ \em kalahs }, data-processing representation of an apron of { \em Kalah }. 
We also used a variable containing all the history of a play, with a 
succession of aprons corresponding to the movements carried out, i.e. 
the first component represents the initial apron and then each component 
point out the apron reached according to each movement. This structure of history enabled us to implement the following functionalities easily: To make preceding blows:
 This option makes it possible to the user to reconsider his last blows. 
At the level of data structure, it is enough to use before last apron
 stored in the history as being the current apron. To make following blows: 
Just as to retrogress, to advance it is enough to seek the aprons which are 
on the right current and to shift in this direction of the number of following
 blows.

\section{ Artificial Intelligence } 
The artificial intelligence is a concept 
about which, generally, everyone intended to speak. The artificial 
intelligence gives the possibility "of thinking" of a machine and, therefore,
 allows him to make decisions. In our work, we used it to be able player 
against the computer and which it can choose his movement. This intelligence
 is called artificial because it is based on calculations of the algorithms.
 
\subsection{Strategy } 
The module { \tt Strategy } is the module which 
manages the artificial intelligence of the computer. It is the most 
significant part of our work. It gives the possibility "of thinking", i.e., 
to choose its next movement "grace" mainly with the algorithm { \tt MiniMax } 
(in concrete terms, an alternative of this algorithm) described below. 
For recall, we present initially the operation of the { \tt traditional MiniMax } then the variations that we have introduce. 

\subsection{Algorithm { \tt MiniMax } } 
The algorithm { \tt MiniMax } is a universal algorithm, 
which is used in the plays with two players, to decide which is the best 
movement to be made, at a given time, starting from the current state. In our 
case one considers the current state as the apron of the game( pawns which 
are in each hole and in the { \em kalahs }), and  the player who has the hand. 
One will use this algorithm preferably to allow the computer to choose his 
movement or to be able to give to human council in the play of the type 
{ \it a player against the computer}. Proposed in 1928, by John Von 
Neumann, this technique leads the computer to review all the possibilities
 for a limited number of blows and to assign a value to them which takes into 
account the benefit for the player who has the hand and for his adversary. 
The best choice being then that which maximizes its benefit (one it calls 
player $MAX$) while minimizing those of sound adversary(player $MIN$). 
This algorithm must be able to return a value which will correspond to the 
movement chosen by the player $MAX$. The basic idea consist to create a 
tree which will have as many levels as that indicated by a value passed in 
parameter. Each node of the tree can have  more than one number of wire 
equal to the number of possible movements. For example this number will be 
six for  { \it Kalah } game  with six holes by player. The algorithm 
{ \tt MiniMax } is a  depth search algorithm, with a limited depth. \\
 It requires to use: 
\begin{itemize} 
\item a generation function of the legal blows starting from a position 
\item a evaluation  function of a position of play 
\end{itemize} 
From a position of the play, the algorithm explores the tree of all  legal 
blows until the required depth. The scores of the tree's sheets are then 
calculated  by the evaluation function.
 A positive score indicates a good position for the player $MAX$ and a 
negative score a bad position for him, therefore a good position 
for the player $MIN$. According to one who plays, the passage of a 
position to another is maximizing (for the player $MAX$) or minimizing for 
the player $MIN$. The players try to play the most advantageous blows for 
themselves. By seeking the best blow for $MAX$, the depth search for 
level 1 will seek to determine the immediate blow which maximizes the score 
of the new position. 

\begin{figure}[htbp]
\begin{center}
\includegraphics[height=5.5cm]{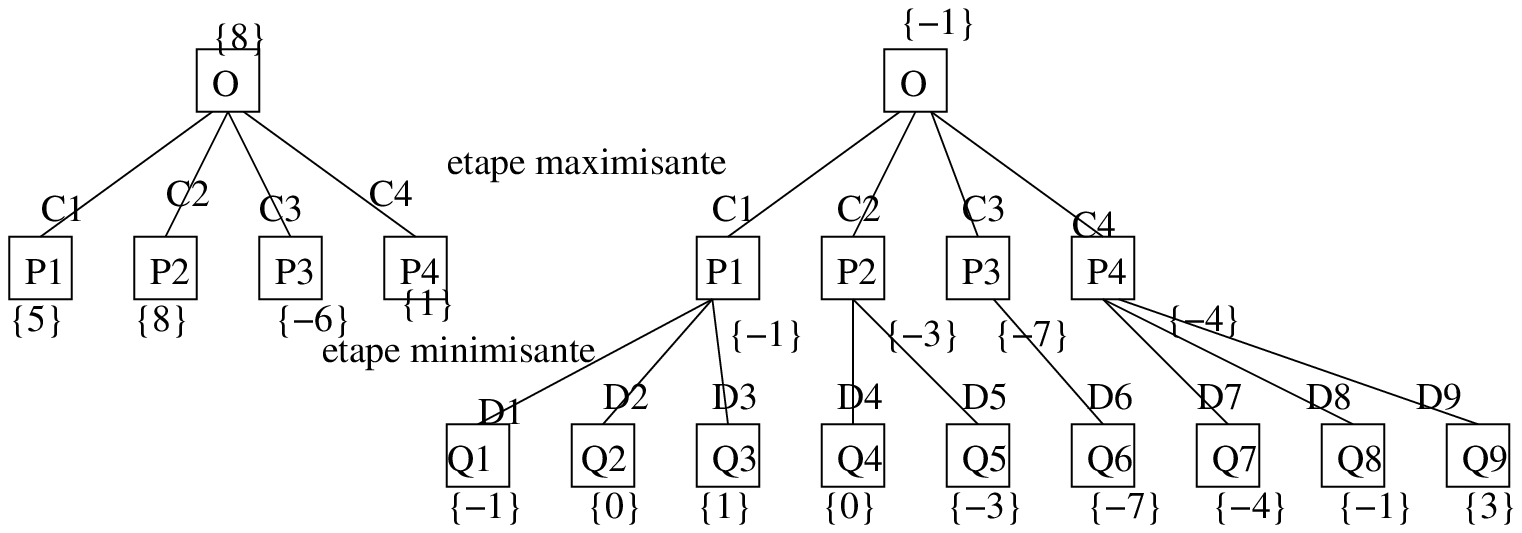}
%\caption{figure1 et figure 2}
\label{maximizing and minimizing steps}
\end{center}
\end{figure}
 For example, on the figure, the player $MAX$ leaves position 0, determines 
four legal blows, builds these new configurations and evaluates them. Of these
 scores, its best position is (of score 8). It propagates this value with 
position 0, indicating speak with this pleasing position in a blow with a new
 position about score 8 by playing the blow $C2$. in-depth exploration about 
level 1 is in general not sufficient, because it does not take account of the 
response of the adversary. That produced of the programs seeking the immediate
 profit (like the catch of a queen to the chess board), without realizing that 
the parts are protected or that the position becomes losing (gambit of the 
queen to make checkmate). An exploration of depth 2 makes it possible to 
realize by-effect. Figure 2 shows an additional level of development of the 
tree by taking account of the answer of the player $MIN$. This one to also 
seek its best blow. For that, the algorithm { \tt MiniMax } will minimize the 
scores of the nodes of depth 2. The blow who brought to a position immediately
 score 8 goes, in made, to indeed bring the position of the play to a score 
of -3. if B plays the blow $D5$, then the score of the position $Q5$ is
 worth -3. We can  see  that the blow $C1$ limits the dégats with a score
 of -1. It will thus be preferred. In the majority of the plays, it is 
possible to make lanterns its adversary, by making it play forced blows, 
with an aim of muddling the situation by hoping that it will make a fault. 
For that the search for depth 2 is very insufficient for the tactical aspect 
of the play. The strategic aspect is seldom well exploited by a program 
because it does not have the vision probable evolution of the position at 
the end of the part. The difficulty larger depth comes from the combinator 
explosion. For example, to the failures, the additional exploration 2 depths 
brings a factor of approximately thousand times of combinations (30*30). 
Therefore, if one seeks to calculate a depth of 10, one will obtain 
approximately 514 position, which is of course too. For that, one tries to 
prune the tree of research to reduce this complexity. 

\subsection{Alphabeta Pruning } 
One can note that it is not forcing useful to explore the branch 
in o\`u measurement the score of this position to depth 1 is already
 with less good than that found in the branch In the same way the branch need does not
 have completely to be explored. As of the calculation of $Q7$, we obtain
 a score lower than that of (always completely explored). Calculates $Q8$ 
and $Q9$ will not be able to improve this situation even if their respective 
score is better than $Q7$. In a minimizing stage, the weakest score went up. 
Already is known that it will not bring anything again. The alternative 
alphabeta of the { \tt MiniMax } uses this pruning to decrease the number of 
branches to be explored. This reduction causes an increase in the performances
 in time and in space. With the alphabeta pruning, we  generate only a number 
of nodes necessary to decide if each branch will bring us to a better value 
of that already  exist. 

\subsection{Algorithm MiniMax Revisited } 
The MiniMax algorithm with alphabeta pruning  is created for games with two 
players in which, after each turn the hand changes. By the specificity of the 
 {\em Kalah} game, there is a rule which 
allows to keep the hand (replay again), then it was necessary to make 
modifications on this algorithm. In the traditional exploration tree, 
one makes the maximization of the wire values to the root 
(one wants the value the greatest value for the player 
$MAX$ who is, by definition, that calls the algorithm. In the following level,
 one makes minimization because it is the turn of the adversary (player $MIN$
. One continues thus while alternating until the end. In the play of Kalah, 
the problem comes from the fact, that by generating the depth tree of game,
 it is necessary to store the information which says if the player who made 
the last movement must replay or not. Alternation is not systematic any more,
 one looks at initially if that which has just played owes replay. In this 
case, if the player is the $MAX$, it will again be necessary to maximize in 
the following level. If not, it will be necessary to minimize. 
This modification implies that alphabeta pruning  cannot be bracket in all the 
cases. It will be  able to apply it only in the following cases: 
\begin{itemize} 
\item One is in a node $MAX$ and his/her father is $MIN$ 
\item One is in a node $MIN$ and his/her father is $MAX$ 
\end{itemize} 
The reason is rather obvious. Indeed, let us consider the example of figure 3, 
First of all, one goes down, in-depth, by the first branch, when we go up, we
 obtain value 2 out of B, then in A, like temporal value. Now, we go
 down by the second branch to the sheet, we go up -1, whatever the values 
which can be obtained they will not exceed -1 is thus will be always lower 
than 2. We can prune these branches, because they will not bring additional information. 

Let us consider the case where  the player $MAX$ keep the hand, and suppose 
that the node C is of type $MAX$. A priori it is not known if one of wire 
of C will not be able to go up a higher value. They will thus have to be 
go through  all.  For example in figure 4, the last wire has value 5 and 
this one is finally the gone up value with A.

With this alternative that appears clear that one loses because each time 
that a player keep the hand, our algorithm generates more nodes and 
consequently, more time. But the nature of {\em Kalah} game does not make 
it possible to optimize more than what we did. 

\subsection{ Levels of  difficulty} 
In the program, we introduced four levels of difficulties which can be chosen 
by the user when it asks a play against the computer. More the selected level 
is  more high, the difficulty increases and more the possibilities of gaining 
decreases. There is a direct correspondence between each one of the levels and
 the size of the  exploration tree generated by the algorithm 
{ \tt MiniMAx }. The easy level corresponds to a tree of depth 2, the mean 
level with 4, the difficult level with 6 and the very difficult level with 8.
 We used a parameter which is used as coefficient. In our case this parameter
 is worth 2. We obtain levels of depth 2,4, 6 and 8. It can be changed, for
 example into 3, and in this case one obtains depths 3, 6, 9 and 12. However,
 it should not be forgotten that the number of nodes generated by the 
algorithm believes in an exponential way in each increase of a unit of level. 
For level 8, with 6 branches(game with 6 holes), the complete tree contains 
more than two billion nodes. For reason of response time of the machine, we 
decided not to go beyond. With eight levels, we obtain an acceptable 
response time.

\subsection{Comparison MiniMax and our algorithm }
 We measure the number of nodes generated by the exploration tree, 
instead of time, because, in this way, we will obtain a measurement which is 
always independent on the computer  which serves  the tests. 
For the algorithm without pruning alphabeta, the number of generated nodes is 
equal $\sum_{i=0}^{n } 6^i $. We tested with various states of the apron and 
various levels. The computed value is the number of nodes generated on 
average. To measure the improvement, we give the percentage of nodes 
generated by our algorithm compared to the traditional one. The improvement 
is very considerable especially with regard to the highest levels. 

\begin{center}
\begin{tabular}{||l|l|l|l||}
\hline

Level & traditional Minimax & Our algorithm & Percentage \\
\hline
 2   &            43  &                                 17,2 &                
         40\%\\

 4   &            1555 &                              226,4    &              
      15\%\\
 6   &            55987   &                          3402,7   &               
    6\%\\
 8 &               2015539   &                     44471,1   &            
     2\%\\
\hline
\end{tabular}
\end{center}
\section{Conclusion } 
We used a universal algorithm, the MiniMax, that one 
uses for the games with two players. It is a question of creating a tree  of a 
limited depth containing all  possible blows for this depth.
In exploring  the tree, the machine can decide its next movement. 
The computing time to generate this tree posed a serious problem. To reduce this computing time we 
implemented a more efficiency version of this algorithm. We limited ourselves 
to the presentation of our MiniMax algorithm without speaking about various 
implemented strategies.

\end{document}